%% file: acl_latex.tex
\pdfoutput=1

\documentclass[11pt]{article}

\usepackage{acl}

\usepackage{times}
\usepackage{latexsym}

\usepackage[T1]{fontenc}

\usepackage[utf8]{inputenc}

\usepackage{microtype}

\usepackage{booktabs}
\usepackage{multirow}
\usepackage{amsmath, amssymb}
\usepackage{tcolorbox}
\usepackage{xspace}
\usepackage{enumitem}

\newcommand{\model}{\textsc{UniST}\xspace}
\newcommand{\basemodel}{\textsc{UniST}\textsubscript{BASE}}
\newcommand{\largemodel}{\textsc{UniST}\textsubscript{LARGE}}
\newcommand{\multimodel}{\textsc{UniST}\textsubscript{U+T+M}}

\newcommand{\stitle}[1]{\vspace{1ex} \noindent{\bf #1}}

\usepackage{cleveref}
\crefformat{section}{\S#2#1#3}
\crefformat{subsection}{\S#2#1#3}
\crefformat{subsubsection}{\S#2#1#3}
\crefrangeformat{section}{\S\S#3#1#4 to~#5#2#6}
\crefmultiformat{section}{\S\S#2#1#3}{ and~#2#1#3}{, #2#1#3}{ and~#2#1#3}
\usepackage{refstyle}
\Crefformat{figure}{#2Fig.~#1#3}
\Crefmultiformat{figure}{Figs.~#2#1#3}{ and~#2#1#3}{, #2#1#3}{ and~#2#1#3}
\Crefformat{table}{#2Tab.~#1#3}
\Crefmultiformat{table}{Tabs.~#2#1#3}{ and~#2#1#3}{, #2#1#3}{ and~#2#1#3}
\Crefformat{appendix}{Appx.~\S#2#1#3}
\crefformat{algorithm}{Alg.~#2#1#3}

\title{Unified Semantic Typing with Meaningful Label Inference}

\author{James Y. Huang,\;~Bangzheng Li\thanks{~~Equal contributions.},\;~Jiashu Xu\footnotemark[1] \and Muhao Chen \\
  University of Southern California \\
   Los Angeles, California, USA \\
  \texttt{\{huangjam,bangzhen,jiashuxu,muhaoche\}@usc.edu} 
  }

\begin{document}
\maketitle
\begin{abstract}

\input{sections/abstract}

\end{abstract}

\section{Introduction}

\input{sections/introduction}

\section{Method}

\input{sections/method}

\section{Experiments}

\input{sections/experiments}

\section{Related Works}

\input{sections/relatedwork}

\section{Conclusion}

\input{sections/conclusion}

\section*{Acknowledgment}

We appreciate the anonymous reviewers for their insightful comments and suggestions. 
This material is supported in part by the DARPA MCS program under Contract No. N660011924033 with the United States Office Of Naval Research, and by the National Science Foundation of United States Grant IIS 2105329.

\bibliography{anthology,custom}
\bibliographystyle{acl_natbib}

\appendix

\input{sections/appendix}

\end{document}

%% file: sections/abstract.tex
Semantic typing aims at classifying tokens or spans of interest in a textual context into semantic categories such as relations, entity types, and event types. The inferred labels of semantic categories meaningfully interpret how machines understand components of text. In this paper, we present \model, a unified framework for semantic typing that captures label semantics by projecting both inputs and labels into a joint semantic embedding space. To formulate different lexical and relational semantic typing tasks as a unified task, we incorporate task descriptions to be jointly encoded with the input, allowing \model to be adapted to different tasks without introducing task-specific model components. \model optimizes a margin ranking loss such that the semantic relatedness of the input and labels is reflected from their embedding similarity. Our experiments demonstrate that \model achieves strong performance across three semantic typing tasks: entity typing, relation classification and event typing. Meanwhile, \model effectively transfers semantic knowledge of labels and substantially improves generalizability on inferring rarely seen and unseen types. In addition, multiple semantic typing tasks can be jointly trained within the unified framework, leading to a single compact multi-tasking model that performs comparably to dedicated single-task models, while offering even better transferability.\footnote{Our code and pre-trained models are available at \url{https://github.com/luka-group/UniST}.}

%% file: sections/introduction.tex

Semantic typing is a group of fundamental natural language understanding problems that aim at classifying tokens (or spans) of interest into semantic categories. This includes a wide range of long-standing NLP problems such as entity typing, relation classification, and event typing. Inferring the types of entities, relations or events mentioned is not only crucial to the structural perception of human language, but also plays an important role in many downstream tasks such as entity linking \cite{onoe-durrett-2020-interpretable}, information extraction \cite{zhong-chen-2021-frustratingly} and question answering \cite{yavuz-etal-2016-improving}.

\begin{figure*}[ht]
    \centering
    \includegraphics[width=\textwidth]{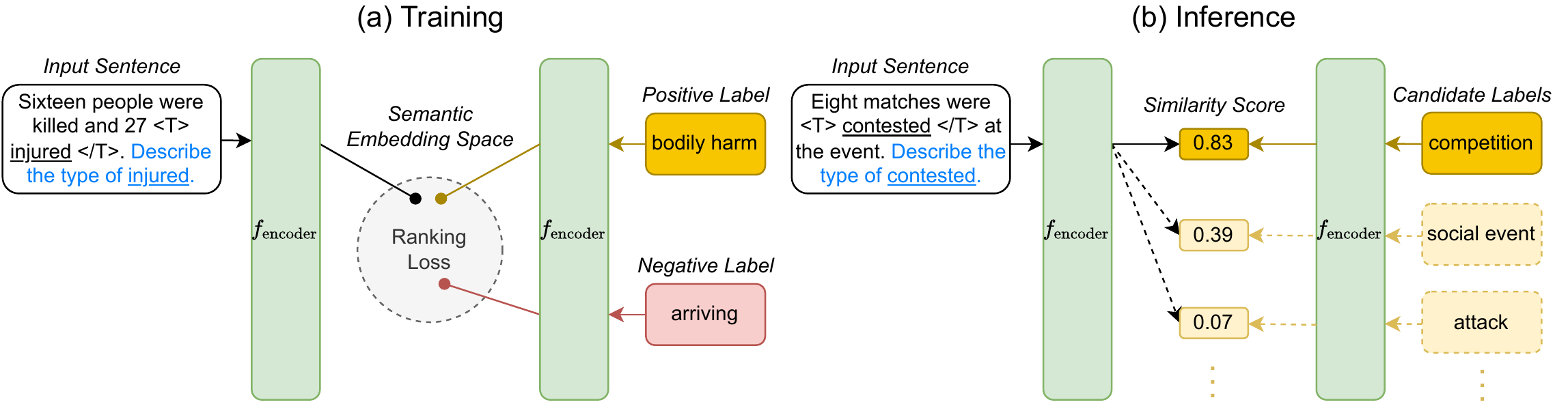}
    \caption{\model projects the input sentence with task descriptions (in blue) and marked token span of interest (with enclosing special tokens), and candidate labels into a shared semantic embedding space. In training, it optimizes a margin ranking loss such that positive labels are closer to the input sentence than negative labels. During inference, \model simply ranks candidate labels based on the similarity between input and label embeddings.}\label{fig:overview}
\end{figure*}

Most traditional methods tackle semantic typing problems by training task-specific multi-class classifiers with token or sentence representations from language models to predict a probability distribution over a pre-defined set of classes \cite{dai-etal-2021-ultra, yamada-etal-2020-luke}. However, this approach comes with several limitations. First, these models simply convert labels into indices, thus completely ignoring the rich semantics carried by the label text itself. For example, given ``\textit{Currently Ritek is the largest producer of OLEDs.}'', knowing what the entity type \textit{company} means would naturally simplify the inference of ``\textit{Ritek}'' is a \textit{company} in this context. Second, models trained as classifiers do not generalize well to class labels that are rarely seen or unseen in the training data, as these models rely on the abundance of annotated examples to associate semantics to label indices. In particular, since these classifiers are limited by the pre-defined label set, they cannot infer any unseen labels unless being re-trained or incorporated with label mapping rules. As a result, these models struggle to handle more fine-grained semantic typing tasks in real-world scenarios \cite{choi-etal-2018-ultra, chen-etal-2020-trying} where any free-form textual labels may be used to represent the types, many of which may also be unseen during training.

In contrast to the aforementioned traditional paradigm for semantic typing, several studies have explored alternative approaches such as prompt-based learning  \cite{schick-schutze-2021-exploiting,ding2021prompt} and indirect supervision from NLI models \cite{yin-etal-2019-benchmarking,sainz-etal-2021-label} to make more efficient use of label semantics. However, these methods usually require hand-crafted templates or mapping between labels and language model vocabulary that do not scale well to diverse, free-form labels across various semantic typing tasks. Instead, we seek a generalizable approach that captures label semantics while requiring minimal effort to be adapted to a different task.

In this paper, we propose \model, a unified framework for semantic typing that projects context sentences and candidate labels into a shared semantic embedding space. \model provides a unified solution to two major categories of semantic typing tasks, namely lexical typing (e.g., entity typing, event typing) and relational typing (relation classification). By optimizing a margin ranking loss, our model captures label semantics such that positive labels are encoded closer to their respective context sentences than negative labels by at least a certain similarity margin. Depending on the task requirement, either top-$k$ candidate labels or any candidate labels with similarity above a certain threshold are given as the final predictions. Furthermore, we add a \textit{task description} to the end of the context sentences to specify the task and token (spans) of interest, and use a single model for encoding both context sentences and labels. This simple technique allows us to unify different semantic typing tasks without introducing separate task-specific model components or learning objectives, while differentiating among distinct task prediction processes during inference. \model demonstrates strong performance on three semantic typing benchmarks: UFET \cite{choi-etal-2018-ultra} for (ultra-fine) entity typing, TACRED \cite{zhang-etal-2017-position} for relation classification, and MAVEN \cite{wang-etal-2020-maven} for event typing, even achieving comparable performance with a single model trained to solve all three tasks simultaneously.

The main contributions of this work are three-fold. First, the proposed \model framework converts distinct semantic typing tasks into a unified formulation, where both input and label semantics can be effectively captured in the same representation space. Second, we incorporate a model-agnostic task representation scheme to allow the model to differentiate among distinct tasks in training and inference without introducing additional task-specific model components. Third, \model demonstrates substantial improvements in both effectiveness and generalizability on entity typing, relation classification and event typing. In addition, our unified framework makes it possible to learn a single model for all three tasks, which performs comparably to dedicated models trained separately on each task.

%% file: sections/method.tex
\input{tables/input}

In this section, we present the technical details of \model, our unified framework for semantic typing. We first provide a general definition of a semantic typing problem (\Cref{probdef}), followed by a detailed description of our model  (\Cref{model}), training objective(\Cref{train}), and inference (\Cref{infer}).

\subsection{Problem Definition} \label{probdef}
Given an input sentence $s$ and a set of one or more token spans of interest $E=\{e_1,...e_n\}, e_i\subset s$, the goal of semantic typing is to assign a set of one or more labels $Y=\{y_1,...y_k\}, Y\subset \mathcal{Y}$ to $E$ that best describes the semantic category $E$ belongs to in the context of $s$. $\mathcal{Y}$ denotes the set of candidate labels, which may include a large number of free-form phrases \cite{choi-etal-2018-ultra} or ontological labels \cite{zhang-etal-2017-position}. In this paper, we consider two categories of semantic typing tasks, \emph{lexical typing} of a single token span (e.g., entity or event typing), and \emph{relational typing} between two token spans (relation classification).

\subsection{Model} \label{model}
\stitle{Overview.} 
As illustrated in \Cref{fig:overview}, \model leverages a pre-trained language model (PLM) to project both input sentences and the candidate labels into a shared semantic embedding space, where the semantic relatedness between the input and label is reflected by their embedding similarity. This is accomplished by optimizing a margin ranking objective that pushes negative labels away from the input sentence while pulling the positive labels towards the input. This simple, unified paradigm allows our model to rank candidate labels based on the affinity of semantic representations with regard to the input during inference. Meanwhile, our model is not limited to a pre-defined label set, as any textual label, whether seen or unseen during training, can be ranked accordingly as long as the model captures its semantic representation. In order to specify the task at hand along with the tokens (or spans) we aim to classify, we add a task description to the end of the input sentence. This allows our framework to use unified representations from a single encoder for both inputs and labels, as well as support the inference of distinct semantic typing tasks without introducing task-specific model components.

\stitle{Task Description.} To highlight the tokens (or spans) we aim to type, we first enclose them with special marker tokens indicating their roles (entities, subjects, objects, or triggers). Next, we leverage the existing semantic knowledge in PLMs and add a natural language task description to the end of the input sentence to specify the task at hand along with tokens (or spans) of interest. The general format for lexical semantic typing is 
\begin{tcolorbox}
\vspace{-0.5em}
\small
        \textit{Describe the type of <tokens>}.
\vspace{-0.5em}
\end{tcolorbox}
\noindent and that of relational semantic typing is
\begin{tcolorbox}
\vspace{-0.5em}
\small
        \textit{Describe the relationship between <subject> and <object>}.
\vspace{-0.5em}
\end{tcolorbox}
\noindent
Examples of different input formats (including special tokens and task descriptions) can be found in \Cref{tab:input}. In addition, relational typing (relation classification) tasks may incorporate entity types from NER models alongside input sentences. Entity type information has been shown to benefit relation classification \cite{peng-etal-2020-learning,zhong-chen-2021-frustratingly,zhou2021improved}, and can be easily incorporated into our task description, as shown in the given example. 

\stitle{Input Representation.}
We use a RoBERTa model \cite{liu2019roberta} to jointly encode the input sentence and the task description. Given an input $s$ and its task description $d$, we concatenate $s$ and $d$ into a single sequence, and obtain the hidden representation of the \texttt{<s>} token as the input sentence representation, denoted by $\mathbf{u}$:
\[\mathbf{u}=f_{\text{encoder}}([s, d]) .\]
A traditional approach to semantic typing is to train classifiers on top of the representations of specific tokens of interest \cite{wang-etal-2021-k,yamada-etal-2020-luke}. In the case of relational typing where two entities are involved, their representations are usually concatenated, leading to dimension mismatch with lexical typing tasks and requiring a different task-specific module to handle.
Instead, thanks to the introduction of task description, \model always uses the universal \texttt{<s>} token representation for both inputs and labels, and across different semantic typing tasks.

\stitle{Label Representation.}
Most semantic typing tasks provide textual labels in natural language from which a language model can directly capture label semantics. Some relation classification datasets such as TACRED use extra identifiers \textit{per:} and \textit{org:} to distinguish same relation type with different subject types. For example, \textit{per:parent} refers to the parent of a person, while \textit{org:parent} represents the parent of an organization such as a company. In this case, we simply replace \textit{per:} and \textit{org:} with \textit{person} and \textit{organization} respectively. The label text is encoded by the exact same model used to encode the input sentence. Given the label $y$, we again take the \texttt{<s>} token representation as the label representation, denoted by $\mathbf{v}$:
\[\mathbf{v}=f_{\text{encoder}}(y) .\]

\subsection{Learning Objective} \label{train}
Let $\mathcal{Y}$ be the set of all candidates labels for a semantic typing task. Given an input $[s,d]$ and the positive label set $Y\subset \mathcal{Y}$, we first randomly sample a negative label $y'\in \mathcal{Y} \backslash Y$ for each training instance. Then, we encode the input $[s,d]$, positive label $y$ and negative label $y'$ into their respective semantic representations $\mathbf{u}$, $\mathbf{v}$, and $\mathbf{v'}$. \model optimizes a margin ranking loss such that positive labels, which are more semantically related to the input than negative labels, are also closer to the input in the embedding space. Specifically, the loss function for a single training instance is defined as:
\[\mathcal{L}_{s,y,y'} = \text{max}\{c(\mathbf{u}, \mathbf{v'})-c(\mathbf{u}, \mathbf{v})+\gamma,0\},\]
where $c(\cdot)$ denotes cosine similarity and $\gamma$ is a non-negative constant. The overall (single-task) training objective is given by:
\[\mathcal{L}_t = \frac{1}{N_t}\sum_{s\in S_t}\sum_{y\in Y_s}\mathcal{L}_{s,y,y'},\]
where $S_t$ is the set of training instances for task $t$, $Y_s$ is the set of all positive labels of $s$, and $N_t$ is the number of distinct pairs of training sentence and positive label. In addition to the single-task setting which optimizes an individual task-specific loss $\mathcal{L}_t$, we also consider a multi-task setting of \model where it is jointly trained on different semantic typing tasks and optimizes the following objective:
\[\mathcal{L} = \frac{1}{N}\sum_{t\in T}\sum_{s\in S_t}\sum_{y\in Y_s}\mathcal{L}_{s,y,y'}.\]
where $T$ is the set of semantic typing tasks \model is trained on, and $N$ is the total number of training instances.

\subsection{Inference} \label{infer}
\model supports different strategies for inference depending on the task requirement. If the number of labels for each input is fixed, we simply retrieve the top-$k$ closest candidate labels to the input as the final predictions. Otherwise, all candidate labels with similarity above a certain threshold are given as predictions. Note that \model is not restricted to a pre-defined label set, as any textual label in natural language can be encoded by \model into its semantic representation and ranked accordingly during inference.

%% file: tables/input.tex
\begin{table*}[t]
    \centering
    \small
    \begin{tabular}{c|c}
        \toprule
        Task & Input Format \\ 
        \midrule
        \multirow{2}{*}{Entity Typing} & Currently <E> \underline{Ritek} </E> is the largest producer of OLEDs. \\ 
        & Describe the type of \underline{Ritek}.\\
        \midrule
        \multirow{2}{*}{Relation Classification} & <SUBJ> \underline{Herrera} </SUBJ> 's wife <OBJ> \underline{Ramona} </OBJ> died in 1991. \\ 
        & Describe the relationship between \underline{\textit{person} Herrera} and \underline{\textit{person} Ramona}.\\
        \midrule
        \multirow{2}{*}{Event Typing} & The siege <T> \underline{began} </T> on 15 September. \\ 
        & Describe the type of \underline{began}.\\ 
        \bottomrule
    \end{tabular}
    \caption{Input formats for different semantic typing tasks. The four pairs of special tokens marks entities, subjects, objects and triggers respectively.}
    \label{tab:input}
\end{table*}

%% file: sections/experiments.tex
In this section, we evaluate \model on single-task experiments on three semantic typing tasks: entity typing (\Cref{entity}), relation classification (\Cref{relation}) and event typing (\Cref{event}). We then assess the generalizability of \model by conducting zero-shot and few-shot prediction, and study the effects of task description (\Cref{analysis}). Finally, we train \model under multi-task setting to solve all three tasks simultaneously (\Cref{multi}).

\subsection{Ultra-fine Entity Typing} \label{entity}
We first conduct experiments on the ultra-fine entity typing task, which aims at predicting fine-grained free-form words or phrases that describe the appropriate types of entities mentioned in sentences.

\stitle{Dataset.}
We use the Ultra-Fine Entity Typing (UFET) benchmark \cite{choi-etal-2018-ultra}, which includes 5,994 sentences split into 1,998 each for train, dev and test. Each entity mention in UFET is annotated with one or more free-form type labels, covering a set of 2,519 distinct words and phrases. Following the original evaluation protocol, we report macro precision, recall and F1 score on the UFET test set.

\stitle{Model.}
Since the number of ground truth labels for each entity is not fixed, all candidate labels with similarity above a certain threshold is given as the final predictions. We tune the hyperparameters, including the threshold, on the UFET dev set. We use base and large versions of RoBERTa as encoders for \basemodel\xspace and \largemodel\xspace respectively.

\stitle{Baselines.} \textbf{UFET-biLSTM} \cite{choi-etal-2018-ultra} learns context and mention representations by combining pre-trained word embeddings with a character-level CNN and a bi-LSTM. \textbf{LabelGCN} \cite{xiong-etal-2019-imposing} adds a graph propagation layer to capture label dependencies. \textbf{LDET} \cite{onoe-durrett-2019-learning} learns a denoising model that automatically filters and relabels distant supervision data for training. \textbf{Box4Types} \cite{onoe-etal-2021-modeling} introduces box embeddings to represent type hierarchies and uses BERT\textsubscript{LARGE} as context and mention encoder. \textbf{LRN} \cite{liu-etal-2021-fine} uses an auto-regressive LSTM to discover label structures, a bipartite attribute graph to capture intrinsic label dependencies, and a BERT\textsubscript{BASE} as sentence encoder. \textbf{MLMET} \cite{dai-etal-2021-ultra} generatively augments the training data with a masked language model, and fine-tunes BERT\textsubscript{BASE} on the augmented training set.

\stitle{Results.}
As shown in \Cref{tab:ufet}, \basemodel\xspace already outperforms the SOTA baseline MLMET without training on any augmented data by 0.2\% in F1 score. With a larger language model, \largemodel\xspace further improves F1 score by another 0.6\%. Since UFET only provides a small set of human annotated training data compared to its diverse label set, all baselines except LRN incorporate distant supervision data to alleviate data scarcity. \model's superior performance on UFET demonstrates the importance of capturing label semantics as an auxiliary supervision signal that is not fully exploited by previous methods. This is especially beneficial when annotated data are limited, and can alleviate the model's reliance on augmenting training data. In this way, \model also achieves better generalizability to unseen and rarely seen labels, for which we conduct a more detailed analysis on few-shot and zero-shot UFET labels in \Cref{analysis}.

\input{tables/ufet}

\subsection{Relation Classification} \label{relation}
The goal of relation classification is to determine the relation between a subject entity and an object entity mentioned in a sentence.

\input{tables/tacred}

\stitle{Dataset.}
We run the experiments on TACRED \cite{zhang-etal-2017-position}, a widely used benchmark for this task that contains 106,264 sentences with entity pairs labeled as one of the 41 relation types or a \textit{no\_relation} type. TACRED provides 68,124 instances for training, 22,631 for dev, and 15,509 for testing. Following the original evaluation protocol, we report micro precision, recall and F1 score on the TACRED test set.

\stitle{Model Configuration.}
\model retrieves the candidate label closest to the input in the embedding space as the final prediction. Since entities in TACRED are also annotated with entity types, we place the entity type labels in front of their corresponding entity mentions in the task description to provide additional information for relation classification, as shown in \Cref{tab:input}. We tune the hyperparameters on the TACRED dev set.

\stitle{Baselines.} \textbf{SpanBERT} \cite{joshi-etal-2020-spanbert} incorporates span prediction as an additional objective for BERT pre-training. \textbf{MTB} \cite{baldini-soares-etal-2019-matching} introduces matching-the-blank training on entity-linked text to connect relation representations among related instances. \textbf{TANL} \cite{paolini2021structured} proposes a unified text-to-text framework for structured prediction tasks based on T5 \cite{2020t5}. \textbf{K-Adapter} \cite{wang-etal-2021-k} learns adapter modules to infuse structured knowledge into a RoBERTa\textsubscript{LARGE} model. \textbf{LUKE} further trains RoBERTa\textsubscript{LARGE} on entity-annotated corpus with an entity-aware self-attention mechanism. \textbf{BERT-CR} \cite{zhou-chen-2021-learning} introduces a co-regularization framework to improve learning from noisy datasets with a BERT\textsubscript{LARGE} model. \textbf{IBRE} \cite{zhou-chen-2021-learning} incorporates entity type information into mention markers in the sentence to boost the performance of RoBERTa\textsubscript{LARGE}. \textbf{SP} \cite{cohen2020relation} formulates relation classification as a two-way span prediction problem, and uses ALBERT \cite{Lan2020ALBERT:} as encoder\footnote{We were unable to reproduce the results of RECENT \cite{lyu-chen-2021-relation} due to an error in its evaluation process that wrongly corrected all false positive predictions during testing. After correcting that error, the performance of RECENT was observed to be below the other baselines, and hence has not been included in the result discussion.}. 


\stitle{Results.} As shown in \Cref{tab:tacred}, \basemodel\xspace already outperforms several strong baselines which are built on larger PLMs (BERT\textsubscript{LARGE} or RoBERTa\textsubscript{LARGE}), except for SP and IBRE. \largemodel\xspace further improves the performance and establishes new SOTA on TACRED, outperforming the best baseline SP by 0.7\% in F1. While SP also leverages label semantics by framing relation classification as a two-way question answering problem, it requires hand-crafted question templates for each relation label and more significant computational cost for answer span prediction. In comparison, \model directly captures label semantics from the label text itself, while offering superior performance and inference efficiency as labels can be retrieved by simply computing embedding cosine similarity.


\subsection{Event Typing} \label{event}
Event typing aims at assigning an event type to an event trigger that clearly indicates an event.

\stitle{Dataset.}
We conduct the evaluation using MAVEN \cite{wang-etal-2020-maven}, a general-domain event extraction benchmark with 77,993/18,904/21,835 event triggers for train/dev/test annotated with 168 distinct event types. MAVEN also provides a large set of negative triggers, which includes all content words (nouns, verbs, adjectives, and adverbs) labeled by a part-of-speech tagger but not annotated as an event trigger. Since \model focuses on semantic typing and does not handle mention span prediction, we train a BERT-CRF model to first identify trigger candidates following \citet{wang-etal-2020-maven}, and then predict an event type for each trigger candidate using \model. Following the original paper, we report micro precision, recall and F1 score on MAVEN test set.  

\input{tables/maven}

\stitle{Model Configuration.}
We retrieve the candidate label with the highest similarity to the input as the predicted event type. We tune the hyperparameters on the MAVEN dev set.  

\stitle{Baselines.} \textbf{DMCNN} \cite{chen-etal-2015-event} uses a CNN with dynamic multi-pooling to obtain trigger representations for classification. \textbf{MOGANED} \cite{yan-etal-2019-event} proposes a multi-order GCN to capture interrelation between event trigger and argument representations based on dependency trees. \textbf{DMBERT} \cite{wang-etal-2019-adversarial-training} improves DMCNN by training a BERT\textsubscript{BASE} model as sentence encoder with dynamic multi-pooling. \textbf{BERT-CRF} stacks a CRF layer on top of BERT\textsubscript{BASE} to model multiple event correlations in a single sentence. \textbf{CLEVE} \cite{wang-etal-2021-cleve} proposes a contrastive learning framework fine-tuned on large-scale corpus with AMR structures obtained from AMR parsers, and combines AMR graph representations from a GNN and text representations from RoBERTa\textsubscript{LARGE} to classify event types.

\stitle{Results.}
As shown in \Cref{tab:maven}, \model is able to improve event typing over BERT-CRF, and outperform all baselines except CLEVE. Note that in addition to being initialized from the same RoBERTa model as \model, CLEVE is further fine-tuned on large-scale corpus with AMR structures obtained from a separate parsing model \cite{xu-etal-2020-improving} that also requires large human-annotated data to train. 
This indicates much more expensive supervision signals used by CLEVE.
In contrast, \model effectively captures the meaning of event types and learns to classify event triggers by only fine-tuning on MAVEN,
while still achieving promising performance without the need of any additional annotated resources.

\subsection{Analysis} \label{analysis}
In this section, we provide a detailed analysis to better understand the generalizability of \model and the effects of incorporating task description. Specifically, we examine \model's performance on few-shot and zero-shot entity typing on UFET, zero-shot relation classification on FewRel \cite{han-etal-2018-fewrel}, and how \model performs without task descriptions.


\input{tables/few}

\stitle{Few-shot \& Zero-shot Entity Typing.} A large portion of UFET test set labels have very few or even no training instances. We focus on entity types with no more than 10 instances in the training set, and compare the performance of \basemodel\xspace with the previous SOTA model MLMET on these few-shot and zero-shot labels.

As shown in \Cref{fig:few}, the advantage of \model over MLMET becomes more evident for rarer labels. For the most challenging zero-shot labels, \model substantially outperforms MLMET by 7.2\% in F1 score, suggesting that \model is better generalized to infer low-resource and unseen entity types.

\stitle{Zero-shot Relation Classification.} We conduct experiments on FewRel \cite{han-etal-2018-fewrel}, a widely used benchmark for low-resource relation classification. FewRel includes 64/16/20 non-overlapping relation types for train/dev/test with 700 sentences collected from Wikipedia for each relation type. We evaluate \model under the $N$-way-$0$-shot setting, where the goal is to predict the correct relation among $N$ candidate relations without seen training examples. Following previous studies \cite{cetoli-2020-exploring,dong-etal-2021-mapre}, we report 5-way-0-shot and 10-way-0-shot accuracy on the FewRel dev set.

We compare \model with following baselines: \textbf{REGRAB} \cite{qu2020few} proposes a bayesian meta-learning method to infer the posterior distribution of relation prototypes initialized with knowledge graph embeddings. \textbf{BERT-SQuAD} \cite{cetoli-2020-exploring} formulates zero-shot relation classification as a question answering problem, and fine-tunes a BERT\textsubscript{LARGE} QA model trained on SQuAD 1.1 \cite{rajpurkar-etal-2016-squad} to predict relation types. \textbf{MapRE} \cite{dong-etal-2021-mapre} proposes a contrastive pre-training framework that learns input and relation representations from large-scale relation-annotated data. All baselines, as well as \model, are fine-tuned on the FewRel training set, and then evaluated on the FewRel dev set with a new set of relation types completely \emph{disjoint} from that of the training set.

As shown in \Cref{tab:fewrel}, \model outperforms the best baseline MapRE by 0.5\% and 1.4\% in accuracy on 5-way-0-shot and 10-way-0-shot tasks without first pre-training on any relation-annotated data. This demonstrates that by effectively captures label semantics, \model allows better knowledge transfer to handle unseen relation types.

\stitle{Effects of Task Description.} We conduct an ablation experiment on task descriptions using \model to better understand their effects on downstream tasks. As shown in \Cref{tab:ablation}, 
the performance on TACRED degrades much more significantly compared to that on UFET and MAVEN after removing task description. In lexical typing, the token span to be classified tend to share similar semantics with its type, and in many cases can be easily matched to its type label without explicitly specifying the task. In contrast, relation types are usually not semantically similar to its subject and object, and task description helps bridge this gap.

\subsection{Multi-task Learning} \label{multi}

\input{tables/fewrel}

\input{tables/ablation}

\input{tables/multitask}

With a unified task formulation, \model facilities learning a single model to jointly train on and simultaneously solve different semantic typing tasks. For more balanced training, We train \model on the combined training set of UFET, TACRED and MAVEN, and report F1 performance on their respective test sets by following their respective evaluation protocol. We also include performance of single-task \model for comparison.

As shown in \Cref{tab:multitask}, our multi-task model  obtain generally comparable performance to dedicated \model models trained separately on each of the three semantic typing tasks. Despite a slight decrease in performance on some of the tasks, \multimodel\xspace is still able to outperform several strong baselines discussed earlier. Hence, \model provides a possible solution for learning a compact, unified model with a joint semantic embedding space across different semantic typing tasks. Moreover, this leads to a well-structured embedding space that better allows zero-shot transfer to new semantic typing tasks. To provide a preliminary analysis on the potential of \model on cross-task transfer, we evaluate both single-task and multi-task \model models on FewRel dev set without training on any FewRel data. While FewRel is also a relation classification dataset like TACRED, 75\% of the relation types in FewRel dev set do not exist in TACRED. Results in \Cref{tab:multitask} show that by jointly training on different semantic typing tasks within a unified framework, \model demonstrate significantly stronger transferability to the unseen FewRel task compared to single-task variants. It would be meaningful to see if incorporating more datasets and tasks into \model would further benefit cross-task transfer, especially to tasks with limited data available for training. We leave this as a direction for further investigation.

%% file: tables/ufet.tex
\begin{table}[t]
    \centering
    \small
    \begin{tabular}{l|ccc}
        \toprule
        Model & P & R & F1 \\
        \midrule
        UFET-biLSTM\textsuperscript{\textdagger} \cite{choi-etal-2018-ultra} & 48.1 & 23.3 & 31.3 \\
        LabelGCN\textsuperscript{\textdagger} \cite{xiong-etal-2019-imposing} & 50.3 & 29.2 & 36.9 \\
        LDET\textsuperscript{\textdagger} \cite{onoe-durrett-2019-learning} & 51.5 & 33.0 & 40.1 \\
        Box4Types*\textsuperscript{\textdagger} \cite{onoe-etal-2021-modeling} & 52.8 & 38.8 & 44.8 \\
        LRN \cite{liu-etal-2021-fine} & \textbf{54.5} & 38.9 & 45.4 \\
        MLMET\textsuperscript{\textdagger} \cite{dai-etal-2021-ultra} & 53.6 & 45.3 & 49.1 \\
        \midrule
        \basemodel & 49.2 & 49.4 & 49.3\\
        \largemodel* & 50.2 & \textbf{49.6} & \textbf{49.9}\\
        \bottomrule
    \end{tabular}
    \caption{Results of entity typing on UFET. * marks models based on large versions of PLMs. \textsuperscript{\textdagger} marks models using augmented training data.}
    \label{tab:ufet}
\end{table}

%% file: tables/tacred.tex
\begin{table}[t]
    \centering
    \small
    \setlength{\tabcolsep}{4pt}
    \begin{tabular}{l|ccc}
        \toprule
        Model & P & R & F1 \\
        \midrule
        SpanBERT* \cite{joshi-etal-2020-spanbert} & 70.8 & 70.9 & 70.8 \\
        MTB* \cite{baldini-soares-etal-2019-matching} & - & - & 71.5 \\
        TANL \cite{paolini2021structured} & - & - & 71.9 \\
        K-Adapter* \cite{wang-etal-2021-k} & 70.1 & 74.0 & 72.0 \\
        LUKE* \cite{yamada-etal-2020-luke} & 70.4 & 75.1 & 72.7 \\
        BERT-CR* \cite{zhou-chen-2021-learning} & - & - & 73.0 \\
        IBRE* \cite{zhou2021improved} & - & - & 74.6 \\
        SP* \cite{cohen2020relation} & 74.6 & \textbf{75.2} & 74.8 \\
        \midrule
        \basemodel & 73.6 & 75.0 & 74.3 \\
        \largemodel* & \textbf{78.0} & 73.1 & \textbf{75.5} \\
        \bottomrule
    \end{tabular}
    \caption{Results of relation classification on TACRED. * marks models based on large versions of PLMs.}
    \label{tab:tacred}
\end{table}

%% file: tables/maven.tex
\begin{table}[t]
    \centering
    \small
    \begin{tabular}{l|ccc}
        \toprule
        Model & P & R & F1 \\
        \midrule
        DMCNN \cite{chen-etal-2015-event} & 66.3 & 55.9 & 60.6 \\
        MOGANED \cite{yan-etal-2019-event} & 63.4 & 64.1 & 63.8 \\
        DMBERT \cite{wang-etal-2019-adversarial-training} & 62.7 & 72.3 & 67.1 \\
        BERT-CRF \cite{wang-etal-2020-maven} & 65.0 & 70.9 & 67.8\\
        CLEVE* \cite{wang-etal-2021-cleve} & 64.9 & \textbf{72.6} & \textbf{68.5} \\
        \midrule
        \basemodel & \textbf{66.7} & 69.9 & 68.3\\
        \largemodel* & 66.5 & 69.7 & 68.1\\
        \bottomrule
    \end{tabular}
    \caption{Results of event typing on MAVEN. * marks models based on large versions of PLMs. All baseline results except CLEVE are taken from \citet{wang-etal-2020-maven}}
    \label{tab:maven}
\end{table}

%% file: tables/few.tex
\begin{figure}[t]
    \centering
    \includegraphics[width=\linewidth]{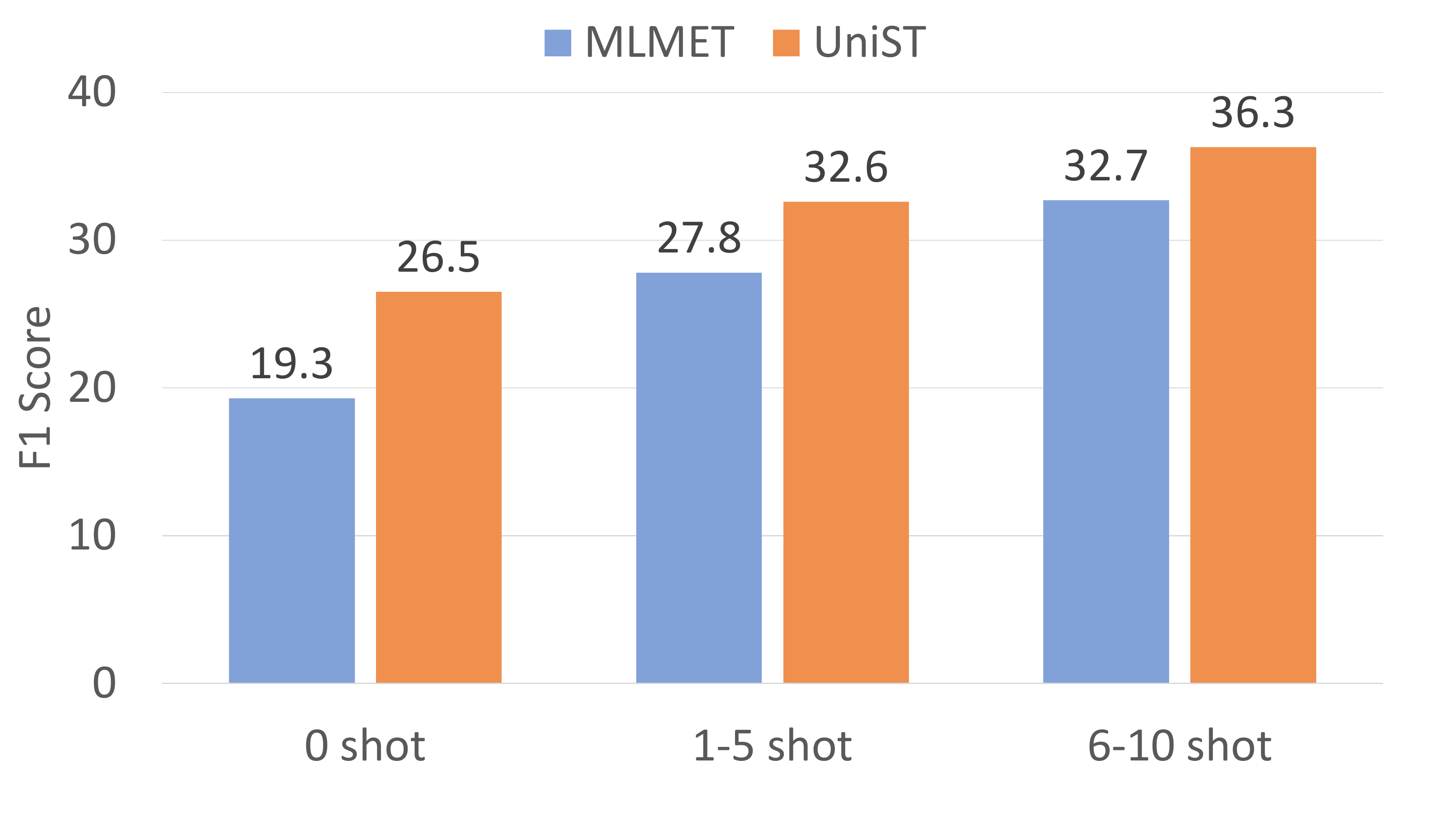}
    \caption{Comparison between MLMET and \model on few-shot and zero-shot prediction on UFET.}
    \label{fig:few}
\end{figure}

%% file: tables/fewrel.tex
\begin{table}[t]
    \centering
    \small
    \begin{tabular}{l|cc}
        \toprule
        \multirow{2}{*}{Model} & 5-way & 10-way \\
         & 0-shot & 0-shot \\
        \midrule
        REGRAB\textsuperscript{\textdagger} \cite{qu2020few} & 52.5 & 37.5 \\
        BERT+SQuAD* \cite{cetoli-2020-exploring} & 86.0 & 76.2 \\
        MapRE \cite{dong-etal-2021-mapre} & 90.7 & 81.5 \\
        \midrule
        \basemodel & 91.2 & 82.9 \\
        \bottomrule
    \end{tabular}
    \caption{Accuracy results of zero-shot relation classification on FewRel. * marks models based on large versions of PLMs. \textsuperscript{\textdagger} Results for REGRAB are taken from \citet{dong-etal-2021-mapre}.}
    \label{tab:fewrel}
\end{table}

%% file: tables/ablation.tex
\begin{table}[t]
    \centering
    \small
    \setlength{\tabcolsep}{4pt}
    \begin{tabular}{l|ccc}
        \toprule
        Model & UFET & TACRED & MAVEN \\
        \midrule
        \basemodel & \textbf{49.3} & \textbf{74.3}  & \textbf{68.3} \\
        - without task description & 49.2 & 72.9 & 68.2 \\
        \bottomrule
    \end{tabular}
    \caption{F1 results of ablation experiments without task description on UFET, TACRED and MAVEN.}
    \label{tab:ablation}
\end{table}

%% file: tables/multitask.tex
\begin{table}[t]
    \centering
    \small
    \setlength{\tabcolsep}{4pt}
    \begin{tabular}{l|ccc|cc}
        \toprule
        \multirow{3}{*}{Model} & \multirow{3}{*}{U} & \multirow{3}{*}{T} & \multirow{3}{*}{M} & \multicolumn{2}{c}{FewRel} \\
        & & & & 5-way & 10-way \\
        & & & & 0-shot & 0-shot \\
        \midrule
        \basemodel\textsubscript{,U} & \textbf{49.3} & 6.1 & 19.6 & 68.9 & 56.0 \\
        \basemodel\textsubscript{,T} & 3.0 & \textbf{74.3} & 5.2 & 62.5 & 48.4 \\
        \basemodel\textsubscript{,M} & 22.5 & 4.3 & \textbf{68.3} & 55.1 & 39.7 \\
        \midrule
        \basemodel\textsubscript{,U+T+M} & 48.5 & 74.2 & 68.2 & \textbf{80.9} & \textbf{72.0} \\
        \bottomrule
    \end{tabular}
    \caption{F1 results by multi-task learning on UFET (U), TACRED (T), MAVEN (M), and zero-shot transfer to FewRel.}
    \label{tab:multitask}
\end{table}

%% file: sections/relatedwork.tex
We present two lines of relevant research topics. Each has a large body of work which we can only provide as a highly selected summary.

\stitle{Semantic Typing.} Semantic typing tasks can be generally categorized into lexical typing (e.g., entity typing, event typing) and relational typing (or classification).  A large number of specialized approaches have been developed for individual semantic typing tasks. For example, prior studies on entity typing have exploited label dependencies and hierarchies \cite{xu-barbosa-2018-neural,xiong-etal-2019-imposing}, capturing label relations with knowledge bases \cite{dai-etal-2019-improving,jin-etal-2019-fine}, as well as automatic data augmentation and denoising techniques \cite{onoe-durrett-2019-learning,dai-etal-2021-ultra} to deal with fine-grained type vocabularies. Relation classification has been tackled by modeling dependency structures \cite{zhang-etal-2018-graph}, learning span representations \cite{joshi-etal-2020-spanbert}, entity representations \cite{yamada-etal-2020-luke}, and injecting external knowledge into pre-trained language models  \cite{peters-etal-2019-knowledge, zhang-etal-2019-ernie, wang-etal-2021-k}. Nevertheless, most previous methods have formulated semantic typing as a multi-class classification problem without capturing label semantics.

\stitle{Learning Label Semantics.} Previous studies have attempted formulating typing tasks into other tasks that allow more effective learning of label semantics. Following this idea, semantic typing tasks have been reformulated as prompt-based learning \cite{ding2021prompt, han2021ptr}, natural language inference \cite{yin-etal-2019-benchmarking, sainz-etal-2021-label}, question answering \cite{levy-etal-2017-zero, li-etal-2019-entity, du-cardie-2020-event}, and translation \cite{paolini2021structured}. Another line of research that is more relevant to our approach focuses on learning semantic label embeddings such that candidate labels can be ranked based on their affinity with the input in the embedding space. Semantic label embeddings have been successfully applied to a variety of tasks such as hierarchical text classification \cite{chen-etal-2021-hierarchy, shen-etal-2021-taxoclass} and intent detection \cite{xia-etal-2018-zero}. In the context of semantic typing tasks, \citet{chen-etal-2020-trying} propose a learning-to-rank framework for multi-axis event process typing with indirect supervision from label glosses. \citet{chen-li-2021-zs} use a pre-trained sentence embedding model to learn relation label embeddings from label descriptions. \citet{dong-etal-2021-mapre} propose a contrastive pre-training framework to learn input and relation representations from large-scale relation-annotated data. Unlike previous approaches, \model does not rely on external label knowledge, training data or task-specific model components. Instead, \model effectively captures label semantics solely from label names, and unify different semantic typing tasks into a single framework by incorporating task descriptions to be jointly encoded with the input.

%% file: sections/conclusion.tex
We propose \model, a unified framework for semantic typing that exploits label semantics to learn a joint semantic embedding space for both inputs and labels. By incorporating model-agnostic task descriptions, \model can be easily adapted to different semantic typing tasks without introducing task-specific model components. Experimental results show that \model offers both strong performance and generalizability on entity typing, relation classification, and event typing. Our unified framework also facilitates learning a single model to solve different semantic typing tasks simultaneously, with performance on par with dedicated models trained on individual tasks.

%% file: sections/appendix.tex
\vspace{2em}
\begin{center}
    {
    \Large\textbf{Appendices}
    }
\end{center}

\section{Experiment Details} \label{appendix}
We run all single-task \basemodel\xspace experiments on NVIDIA RTX 2080Ti GPUs, and all \largemodel\xspace and multi-task experiments on NVIDIA RTX A5000 GPUs. \basemodel\xspace and \largemodel\xspace use base and large versions of RoBERTa as encoders with 125M and 355M parameters respectively. We conduct hyperparameter search within the following range:
\begin{itemize}[leftmargin=*]
\setlength\itemsep{-0.1em}
    \item learning rate: \{3e-6, 5e-6, 1e-5, 2e-5\}
    \item Batch size: \{32, 64, 128\}
    \item Number of training epochs: \{50, 100, 200, 500, 1000\}
    \item Ranking loss margin $\gamma$:\{ 0.1, 0.2, 0.3\}
\end{itemize}
\input{tables/hyp/comhyp}
We optimize our models using AdamW \cite{loshchilov2018decoupled} with linear learning rate decay. The best model checkpoints are selected based on dev set performance. \Cref{tab:comhyp} lists common hyperparameters used across all experiments. All datasets used in our experiments are in English. More details of individual tasks and experiments are provided below.

\subsection{UFET}
The UFET dataset is publicly available on its official website \footnote{\url{https://www.cs.utexas.edu/~eunsol/html_pages/open_entity.html}}. \Cref{tab:ufethyp} shows the hyperparameters and dev F1 score for UFET experiments.
\input{tables/hyp/ufethyp}

\subsection{TACRED}
The TACRED dataset we use is licensed by LDC \footnote{\url{https://catalog.ldc.upenn.edu/LDC2018T24}}. \Cref{tab:tacredhyp} shows the hyperparameters and dev F1 score for TACRED experiments.
\input{tables/hyp/tacredhyp}

\subsection{MAVEN}
The MAVEN dataset is publicly available via its official github repository \footnote{\url{https://github.com/THU-KEG/MAVEN-dataset}}. \Cref{tab:mavenhyp} shows the hyperparameters and dev F1 score for MAVEN experiments.
\input{tables/hyp/mavenhyp}

\subsection{FewRel}
The FewRel dataset is publicly available via its official github repository \footnote{\url{https://github.com/thunlp/FewRel}}. We report the average accuracy of 10 runs on the dev set during evaluation. \Cref{tab:fewrelhyp} shows the hyperparameters for FewRel experiments.
\input{tables/hyp/fewrelhyp}

\subsection{Multi-task Experiments}
We conduct multi-task experiments on the combined UFET, TACRED, and MAVEN training sets. We up-sample UFET training set by a factor of 10 for more balanced training. \Cref{tab:multihyp} shows the hyperparameters and dev set F1 for multi-task experiments.
\input{tables/hyp/multihyp}

\section{Ethics Considerations}
Our experiments are all conducted on openly available and widely used datasets. We do not augment any information to those data in this research, hence this research is not expected to introduce any additional biased information to existing information in those data. However, the model may potentially capture biases reflective of the pre-trained language models and datasets we use for our experiments, in such biases have pre-existed in these pre-trained models or datasets. This is a common problem for models trained on large-scale data, and therefore we suggest conducting a thorough bias analysis before deploying our model in any real-world applications.

%% file: tables/hyp/comhyp.tex
\begin{table}[ht]
    \centering
    \small
    \begin{tabular}{l|c}
        \toprule
        Learning rate & 5e-6 \\
        Dropout rate & 0.1 \\
        Adam $\epsilon$ & 1e-6 \\
        Adam $\beta_1$ & 0.9 \\
        Adam $\beta_2$ & 0.999 \\
        Gradient clipping & 1.0 \\
        Warmup ratio & 0.1 \\
        Ranking loss margin & 0.1 \\
        \bottomrule
    \end{tabular}
    \caption{Common hyperparameters used in all experiments.} 
    \label{tab:comhyp}
\end{table}

%% file: tables/hyp/ufethyp.tex
\begin{table}[h]
    \centering
    \small
    \begin{tabular}{l|cc}
        \toprule
        Name & \basemodel & \largemodel \\
        \midrule
        Batch size & 64 & 64 \\
        Number of training epochs & 1000 & 1000 \\
        Dev F1 & 49.2 & 49.5 \\
        \bottomrule
    \end{tabular}
    \caption{Hyperparameters and dev F1 score for UFET experiments.} 
    \label{tab:ufethyp}
\end{table}

%% file: tables/hyp/tacredhyp.tex
\begin{table}[h]
    \centering
    \small
    \begin{tabular}{l|cc}
        \toprule
        Name & \basemodel & \largemodel \\
        \midrule
        Batch size & 64 & 64 \\
        Number of training epochs & 100 & 100 \\
        Dev F1 & 73.9 & 75.3 \\
        \bottomrule
    \end{tabular}
    \caption{Hyperparameters and dev F1 score for TACRED experiments.} 
    \label{tab:tacredhyp}
\end{table}

%% file: tables/hyp/mavenhyp.tex
\begin{table}[h]
    \centering
    \small
    \begin{tabular}{l|cc}
        \toprule
        Name & \basemodel & \largemodel \\
        \midrule
        Batch size & 64 & 64 \\
        Number of training epochs & 100 & 100 \\
        Dev F1 & 68.4 & 68.5 \\
        \bottomrule
    \end{tabular}
    \caption{Hyperparameters and dev F1 score for MAVEN experiments.} 
    \label{tab:mavenhyp}
\end{table}

%% file: tables/hyp/fewrelhyp.tex
\begin{table}[h]
    \centering
    \small
    \begin{tabular}{l|c}
        \toprule
        Name & \basemodel \\
        \midrule
        Batch size & 64 \\
        Number of training epochs & 50 \\
        \bottomrule
    \end{tabular}
    \caption{Hyperparameters for FewRel experiments.} 
    \label{tab:fewrelhyp}
\end{table}

%% file: tables/hyp/multihyp.tex
\begin{table}[h]
    \centering
    \small
    \begin{tabular}{l|c}
        \toprule
        Name & \basemodel \\
        \midrule
        Batch size & 128 \\
        Number of training epochs & 100 \\
        \multirow{3}{*}{Dev F1} & 47.5 (UFET) \\
        & 72.9 (TACRED) \\
        & 68.3 (MAVEN) \\
        \bottomrule
    \end{tabular}
    \caption{Hyperparameters and dev F1 score for multi-task experiments.} 
    \label{tab:multihyp}
\end{table}